\documentclass{article}

\usepackage[preprint]{neurips_2026}

\usepackage[utf8]{inputenc} 
\usepackage[T1]{fontenc}    
\usepackage{hyperref}       
\usepackage{url}            
\usepackage{booktabs}       
\usepackage{amsfonts}       
\usepackage{nicefrac}       
\usepackage{microtype}      
\usepackage{xcolor}         
\usepackage{graphicx}
\usepackage{amsmath}
\usepackage{pifont}
\usepackage{wrapfig}
\newcommand{\cmark}{\ding{51}}
\newcommand{\xmark}{\ding{55}}

\newtheorem{assumption}{Assumption}
\newtheorem{theorem}{Theorem}
\newtheorem{lemma}[theorem]{Lemma}
\newtheorem{corollary}[theorem]{Corollary}

\newtheorem{proposition}[theorem]{Proposition}

\title{Curvature-Aligned Probing for Local Loss-Landscape Stabilization}

\author{%
  Nikita Kiselev, Andrey Grabovoy \\
  Department of Intelligent Systems \\
  Moscow Institute of Physics and Technology \\
  \texttt{\{kiselev.ns,grabovoy.av\}@phystech.edu}
}

\begin{document}

\maketitle

\begin{abstract}
    Local loss-landscape stabilization under sample growth is typically measured either pointwise or through isotropic averaging in the full parameter space. Despite practical value, both choices probe directions that contribute little to the dominant local deformation of strongly anisotropic neural landscapes. We recast stabilization as an observational problem and introduce a unified family of criteria parameterized by an aggregation order and a probing distribution; within this family we propose a curvature-aligned criterion $\Delta_2^{(D)}$ that probes the loss increment field in the top-$D$ eigenspace of the empirical Hessian near a trained solution. Solely from a local quadratic model, we prove that $\Delta_2^{(D)}$ preserves the $\mathcal O(k^{-2})$ mean-squared rate of the full-space criterion while replacing ambient-dimension curvature dependence with dependence on the subspace dimension $D$; a corollary gives a closed-form spectral expression and a proposition identifies the top-$D$ eigenspace as extremal within the eigenspace-aligned family. We also derive scalable estimators based on Hessian--vector products, subspace Monte Carlo, and a closed-form Gaussian-moment proxy. On a decoder-only transformer, a curvature-aligned probe occupying a tiny fraction of parameter space already reproduces the full-space mean-squared signal to within numerical noise throughout the validated local regime, and the closed-form estimator is orders of magnitude faster than direct Monte Carlo after subspace construction.
\end{abstract}

\section{Introduction}\label{sec:intro}

\begin{figure}[t]
  \centering
  \begin{minipage}[t]{0.48\linewidth}
    \centering
    \includegraphics[width=\linewidth]{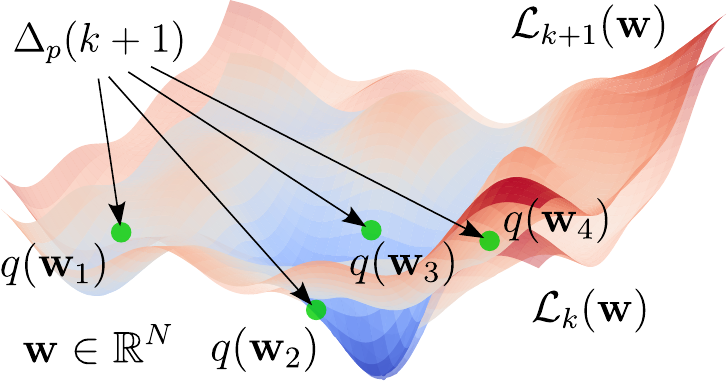}\\[0.3em]
    \small (a) Unified stabilization criterion $\Delta_p$ with probing distribution $q(\mathbf{w})$
  \end{minipage}\hfill
  \begin{minipage}[t]{0.48\linewidth}
    \centering
    \includegraphics[width=\linewidth]{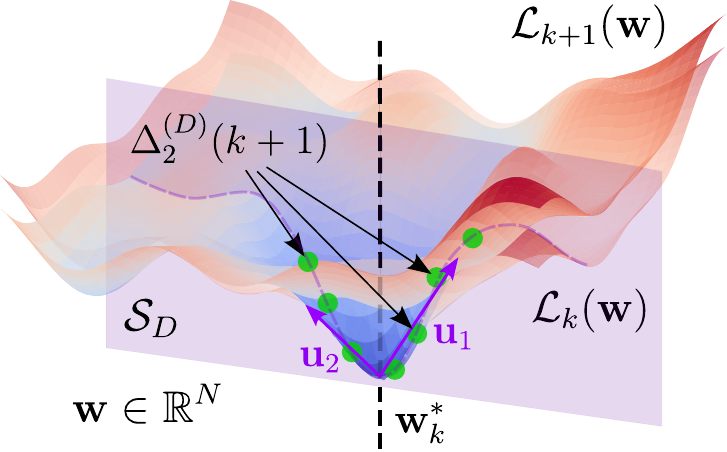}\\[0.3em]
    \small (b) Curvature-aware subspace criterion $\Delta_2^{(D)}$
  \end{minipage}
  \caption{\textbf{Local stabilization as an observational problem.} Existing local criteria differ not only in aggregation order, but also in how they probe the increment field. Our criterion $\Delta_2^{(D)}$ restricts probing to the principal Hessian subspace spanned by the top-$D$ curvature directions.}
  \label{fig:overview}
\end{figure}

Local loss geometry is often summarized through quantities such as sharpness, curvature, or Hessian spectra, typically for a fixed trained model. Our setting is different: we ask how the empirical objective deforms locally as the training set grows. In this regime, the issue is not only which functional of the loss increment should be aggregated, but also which perturbation directions should be used to observe it. Conventional wisdom holds that local criteria should probe parameter space isotropically to avoid bias, or collapse to a single point for tractability~\citep{kiselev2024unraveling}. We argue that neither choice is necessary, and view local stabilization under sample growth as an observational problem in which the probing law is an explicit design variable (Figure~\ref{fig:overview}). Concretely, this lets us ask a quantitative question that the pointwise and isotropic criteria cannot resolve: \emph{how much of one-sample landscape deformation is concentrated in the dominant curvature modes?}

This issue is especially relevant in deep networks, where local geometry is strongly anisotropic. Empirical Hessian spectra typically concentrate much of their mass in a relatively small number of dominant directions, while much of parameter space remains weakly curved or nearly flat~\citep{li2018visualizing,sagun2017hessian,ghorbani2019investigation,papyan2019fullspectrum,xu2024mining}. Related work also suggests that optimization often occupies low-dimensional effective subspaces and that overparameterized models contain substantial geometric redundancy through flat or symmetry-induced directions~\citep{gurari2018tiny,li2018intrinsic,simsek2021geometry,draxler2018essentially,garipov2018mode}. These observations suggest that \emph{how} one probes local deformation should be part of the definition of stabilization, not merely part of the estimator.

We study stabilization under one-sample growth. The object of interest is the increment field $\mathcal L_{k+1}(\mathbf w)-\mathcal L_k(\mathbf w)$ near a trained solution $\mathbf w_k^*$. This places our work next to, but distinct from, classical sample-growth viewpoints. Statistical learning theory and algorithmic stability study convergence of empirical quantities, predictors, or generalization error under changes in the training set~\citep{vapnik1998statistical,shalev2014understanding,bousquet2002stability,hardt2016train,bousquet2020sharper}. Influence-function and infinitesimal-jackknife methods go one step closer by linking data perturbations to local first- and second-order structure~\citep{koh2017influence,giordano2019swiss,koh2019group,basu2020secondorder}. Our target is different: we study the local deformation of the \emph{empirical objective itself}.

We formalize this viewpoint through a unified family of local stabilization criteria parameterized by an aggregation order and a probing distribution; see Eq.~\eqref{eq:general-p-criterion}. This places previously studied one-point and isotropic mean-squared criteria into a common framework and makes the probing law part of the observable. We then propose a curvature-aware specialization, the subspace mean-squared criterion $\Delta_2^{(D)}$ in Eq.~\eqref{eq:delta2-subspace-method}, which restricts probing to the top-$D$ eigenspace of the empirical Hessian at $\mathbf w_k^*$.

Our main result shows that this geometric restriction does not incur a rate penalty under a local quadratic model. Theorem~\ref{thm:subspace-rate} proves that $\Delta_2^{(D)}$ preserves the $\mathcal O(k^{-2})$ mean-squared decay of the full-space criterion, while replacing ambient-dimension curvature dependence by dependence on the probing dimension $D$. In a stable-principal-directions regime, Appendix~\ref{app:proofs} further gives a spectral closed form and an extremality result for the top-$D$ choice within the eigenspace-aligned family.

We also derive scalable estimators based on Hessian--vector products, subspace Monte Carlo, and a closed-form Gaussian-moment proxy. Empirically, we study how the proposed criteria decay with sample size, when the subspace criterion preserves the full-space mean-squared signal, for which perturbation scales the quadratic proxy is accurate, and how the three estimators trade fidelity against efficiency.

\paragraph{Contributions.}
Our contributions are four-fold:
\begin{itemize}
    \item We recast local loss-landscape stabilization under sample growth as an observational problem and introduce a unified family of criteria parameterized by an aggregation order and a probing distribution.
    \item We propose a curvature-aware subspace criterion $\Delta_2^{(D)}$ that probes the increment field in the top-$D$ eigenspace of the empirical Hessian near a trained solution.
    \item Under a local quadratic model, we prove that reducing the probe from $\mathbb R^N$ to a $D$-dimensional curvature-aligned subspace preserves the $\mathcal O(k^{-2})$ mean-squared rate and replaces an $N$-dependent curvature constant by a $D$-dependent one (Theorem~\ref{thm:subspace-rate}).
    \item We develop scalable estimators for the criterion and show empirically that the quadratic proxy is sufficient in its validated local regime and orders of magnitude faster than direct Monte Carlo once the subspace has been constructed.
\end{itemize}

\section{Related Work}\label{sec:related}

\paragraph{Loss geometry and anisotropic local structure.}
A large literature studies neural loss landscapes through visualization, Hessian spectra, sharpness, and related curvature diagnostics. A recurring empirical finding is strong anisotropy: a relatively small number of eigendirections often account for much of the local second-order structure, while large parts of parameter space remain weakly curved or nearly flat~\citep{li2018visualizing,sagun2017hessian,ghorbani2019investigation,papyan2019fullspectrum,xu2024mining}. Related work further suggests that optimization can concentrate in low-dimensional subspaces and that overparameterized models exhibit geometric redundancy through symmetry or flat directions~\citep{gurari2018tiny,li2018intrinsic,simsek2021geometry,draxler2018essentially,garipov2018mode}. These works motivate geometry-aware probing, but they do not study sample-growth deformation of the empirical objective.

\paragraph{Data perturbation, stability, and influence.}
Classical statistical learning theory and algorithmic stability study how empirical risks, predictors, and generalization error behave as the training sample changes~\citep{vapnik1998statistical,shalev2014understanding,bousquet2002stability,hardt2016train,feldman2019high,bousquet2020sharper}. A closer neighboring literature analyzes infinitesimal or finite data perturbations through influence functions and infinitesimal-jackknife approximations, linking changes in the training set to local gradient and Hessian information~\citep{koh2017influence,giordano2019swiss,koh2019group,basu2020secondorder}. These are the closest conceptual precursors to our sample-growth viewpoint. The key difference is the observable: prior work typically measures the sensitivity of fitted parameters or predictions, whereas we measure the local deformation of the empirical objective itself through the increment field $\mathcal L_{k+1}(\mathbf w)-\mathcal L_k(\mathbf w)$.

\paragraph{Sharpness, dominant eigenspaces, and second-order methods.}
Related work also studies flatness, sharpness, sharpness-aware optimization, and dominant Hessian eigenspaces~\citep{keskar2017largebatch,dinh2017sharp,foret2021sam,dauphin2024neglected,luo2024eigensam,singh2021analytic,wu2020dissecting}. These works ask which curvature directions matter for optimization or robustness on a fixed objective. Our question is different: given anisotropic local geometry, which probing law should be used to observe \emph{dataset-induced} deformation? On the computational side, our estimators rely on standard scalable second-order primitives---Hessian--vector products and iterative eigensolvers---rather than explicit Hessian construction~\citep{pearlmutter1994fast,yao2020pyhessian}. Thus, our contribution is not a new second-order tool, but a geometry-aware observable for sample-growth stabilization.

\begin{table}[t]
\caption{Positioning relative to adjacent directions. \cmark~= central focus; \xmark~= not a primary focus.}
\label{tab:positioning}
\begin{center}
\small
\begin{tabular}{p{4.4cm}ccc}
\toprule
\textbf{Direction} & \textbf{Data perturbation} & \textbf{Local objective} & \textbf{Geometric probe} \\
\midrule
Statistical learning theory~\citenum{vapnik1998statistical,shalev2014understanding} & \cmark & \xmark & \xmark \\
Algorithmic stability~\citenum{bousquet2002stability,hardt2016train,feldman2019high,bousquet2020sharper} & \cmark & \xmark & \xmark \\
Influence / infinitesimal jackknife~\citenum{koh2017influence,giordano2019swiss,koh2019group,basu2020secondorder} & \cmark & \xmark & \xmark \\
Sharpness / SAM~\citenum{keskar2017largebatch,dinh2017sharp,foret2021sam,dauphin2024neglected,luo2024eigensam} & \xmark & \xmark & \cmark \\
Hessian spectrum / subspace analysis~\citenum{li2018visualizing,sagun2017hessian,ghorbani2019investigation,papyan2019fullspectrum,xu2024mining,gurari2018tiny,li2018intrinsic,singh2021analytic,wu2020dissecting} & \xmark & \xmark & \cmark \\
Prior increment criteria ($\Delta_1$, $\Delta_2$)~\citenum{kiselev2024unraveling} & \cmark & \cmark & isotropic only \\
\textbf{This work} & \cmark & \cmark & \cmark \\
\bottomrule
\end{tabular}
\end{center}
\end{table}

\section{Method}
\label{sec:method}

\paragraph{Notation.}
Let $\mathfrak{D}_m = \{(\mathbf{x}_i, \mathbf{y}_i)\}_{i=1}^m$ be a training sample of size $m$, and let $\mathcal{L}_m(\mathbf w) = \frac{1}{m}\sum_{i=1}^m \ell\bigl(f_{\mathbf w}(\mathbf x_i), \mathbf y_i\bigr) = \frac{1}{m}\sum_{i=1}^m \ell_i(\mathbf w)$ be the empirical risk of a parametric model, i.e., a neural network $f_{\mathbf w}$, with minimizer $\mathbf w_m^* \in \arg\min_{\mathbf w\in\mathbb R^N} \mathcal L_m(\mathbf w)$. Throughout, subscripts denote per-sample quantities ($\mathbf g_i := \nabla \ell_i$, $\mathbf H_i := \nabla^2 \ell_i$), while parenthesized superscripts denote empirical averages ($\mathbf g^{(m)} := \nabla \mathcal L_m = \frac1m\sum_{i=1}^m \mathbf g_i$, $\mathbf H^{(m)} := \nabla^2 \mathcal L_m$). Our object of interest is the one-sample increment field $\mathcal L_{k+1}(\mathbf w)-\mathcal L_k(\mathbf w)$, which measures how the empirical landscape changes when one example is added. Since this increment is a scalar field over parameter space, any local notion of stabilization depends both on what is aggregated and on how the field is probed.

\paragraph{Unified family of stabilization criteria.}
For $p\ge 1$, we define a general criterion
\begin{equation}
\label{eq:general-p-criterion}
\Delta_p(k+1)
=
\int_{\mathbb R^N}
\bigl|
\mathcal L_{k+1}(\mathbf w)-\mathcal L_k(\mathbf w)
\bigr|^p
\, q(\mathbf w)\, d\mathbf w,
\end{equation}
where $q$ is a probing distribution concentrated near $\mathbf w_k^*$. This definition makes the probing law part of the criterion itself rather than part of the estimator. Previously studied criteria arise as special cases:
\begin{align}
\Delta_1(k+1)
&=
\bigl|
\mathcal L_{k+1}(\mathbf w_k^*)-\mathcal L_k(\mathbf w_k^*)
\bigr|,
&&
q=\delta_{\mathbf w_k^*},
\label{eq:delta1-special}
\\
\Delta_2(k+1)
&=
\int_{\mathbb R^N}
\bigl(
\mathcal L_{k+1}(\mathbf w)-\mathcal L_k(\mathbf w)
\bigr)^2
q(\mathbf w)\,d\mathbf w,
&&
q=\mathcal N(\mathbf w_k^*,\sigma^2\mathbf I_N).
\label{eq:delta2-special}
\end{align}

\paragraph{Curvature-aligned subspace probe.}
In strongly anisotropic models, isotropic perturbations spread mass across many directions that contribute little to the dominant local second-order structure~\citep{sagun2017hessian,ghorbani2019investigation,papyan2019fullspectrum,gurari2018tiny}. We therefore align the probe with the leading curvature directions at $\mathbf w_k^*$. Let $\mathbf H^{(k)}(\mathbf w_k^*) = \mathbf U \mathbf \Lambda \mathbf U^\top$, $\lambda_1^{(k)} \ge \cdots \ge \lambda_N^{(k)}$, and let $\mathbf U_D=[\mathbf u_1,\dots,\mathbf u_D]$ denote the top-$D$ eigenvectors, with principal curvature subspace $\mathcal S_D=\operatorname{Im}(\mathbf U_D)$. We then define the \emph{subspace mean-squared criterion}
\begin{equation}
\label{eq:delta2-subspace-method}
\Delta_2^{(D)}(k+1)
=
\int_{\mathbf w_k^*+\mathcal S_D}
\bigl(
\mathcal L_{k+1}(\mathbf w)-\mathcal L_k(\mathbf w)
\bigr)^2
\, q(\mathbf w)\, d\mathbf w,
\end{equation}
with the parameterization $\mathbf w=\mathbf w_k^*+\mathbf U_D\mathbf z$, where $\mathbf z\sim\mathcal N(\mathbf 0,\sigma^2\mathbf I_D)$. The distinction between $\Delta_2$ and $\Delta_2^{(D)}$ is therefore not merely dimensional: $\Delta_2$ probes the increment isotropically in $\mathbb R^N$, whereas $\Delta_2^{(D)}$ probes it through the dominant local curvature modes of the empirical landscape.

\section{Theoretical analysis}
\label{sec:theory}

Our central theoretical finding is that geometric compression is free in rate: under the local quadratic model in Eqs.~\eqref{eq:taylor-quadratic}--\eqref{eq:loss-diff-minimum}, the subspace criterion $\Delta_2^{(D)}$ preserves the $\mathcal O(k^{-2})$ decay of the full-space criterion while replacing ambient-dimension curvature dependence by dependence on the probing dimension $D$ (Theorem~\ref{thm:subspace-rate}). Within the eigenspace-aligned family, the top-$D$ choice is moreover extremal among all $D$-dimensional eigenspace-aligned probes (Proposition~\ref{prop:extremal-topD}).

\paragraph{Exact increment and local quadratic model.}
Adding one training example to $\mathfrak D_k$ gives the identity
\begin{equation}
\label{eq:loss-increment-exact}
\mathcal L_{k+1}(\mathbf w)-\mathcal L_k(\mathbf w)
=
\frac{1}{k+1}\Bigl(\ell_{k+1}(\mathbf w)-\mathcal L_k(\mathbf w)\Bigr).
\end{equation}
This factor $(k+1)^{-1}$ is the source of the $\mathcal O(k^{-2})$ scaling for squared criteria. Fix $\mathbf w_0\in\mathbb R^N$ and assume that $\mathcal L_k$ and $\mathcal L_{k+1}$ are twice continuously differentiable near $\mathbf w_0$. Their second-order Taylor expansions give
\begin{equation}
\label{eq:taylor-quadratic}
\mathcal L_m(\mathbf w)
\approx
\mathcal L_m(\mathbf w_0)
+
\mathbf g^{(m)}(\mathbf w_0)^\top(\mathbf w-\mathbf w_0)
+
\frac12(\mathbf w-\mathbf w_0)^\top \mathbf H^{(m)}(\mathbf w_0)(\mathbf w-\mathbf w_0),
\end{equation}
where $\mathbf g^{(m)}=\nabla \mathcal L_m$ and $\mathbf H^{(m)}=\nabla^2 \mathcal L_m$. Subtracting the two expansions yields a quadratic model for the increment field:
\begin{multline}
\label{eq:loss-diff-second}
\mathcal L_{k+1}(\mathbf w)-\mathcal L_k(\mathbf w)
\approx
\mathcal L_{k+1}(\mathbf w_0)-\mathcal L_k(\mathbf w_0)
+
\bigl(\mathbf g^{(k+1)}(\mathbf w_0)-\mathbf g^{(k)}(\mathbf w_0)\bigr)^\top(\mathbf w-\mathbf w_0)
\\
+
\frac12(\mathbf w-\mathbf w_0)^\top
\bigl(\mathbf H^{(k+1)}(\mathbf w_0)-\mathbf H^{(k)}(\mathbf w_0)\bigr)
(\mathbf w-\mathbf w_0).
\end{multline}

We now set $\mathbf w_0=\mathbf w_k^*$, where $\mathbf w_k^*$ is a local minimizer of $\mathcal L_k$. Since $\mathbf g^{(k)}(\mathbf w_k^*)=\mathbf 0$, the increment simplifies to
\begin{equation}
\label{eq:loss-diff-minimum}
\mathcal L_{k+1}(\mathbf w)-\mathcal L_k(\mathbf w)
\approx
a_k
+
\mathbf g^{(k+1)}(\mathbf w_k^*)^\top(\mathbf w-\mathbf w_k^*)
+
\frac12(\mathbf w-\mathbf w_k^*)^\top
\mathbf A_k
(\mathbf w-\mathbf w_k^*),
\end{equation}
where $a_k=\mathcal L_{k+1}(\mathbf w_k^*)-\mathcal L_k(\mathbf w_k^*)$, $\mathbf A_k=\mathbf H^{(k+1)}(\mathbf w_k^*)-\mathbf H^{(k)}(\mathbf w_k^*)$.

\paragraph{Subspace reduction.}
Restricting the probe to a $D$-dimensional subspace turns the increment into a low-dimensional quadratic form. In particular, for the principal curvature subspace $\mathcal S_D=\operatorname{Im}(\mathbf U_D)$ with parameterization $\mathbf w=\mathbf w_k^*+\mathbf U_D\mathbf z$, the criterion depends only on the compressed quantities
\[
\mathbf c_k=\mathbf U_D^\top\mathbf g^{(k+1)}(\mathbf w_k^*),
\qquad
\mathbf B_k=\mathbf U_D^\top\mathbf A_k\mathbf U_D.
\]

\begin{lemma}[Reduction on the principal curvature subspace]
\label{lem:subspace-reduction}
Suppose $\operatorname{supp}(q)\subset (\mathbf w_k^*+\mathcal S_D)\cap \mathcal U_R(\mathbf w_k^*)$. Under the parameterization $\mathbf w=\mathbf w_k^*+\mathbf U_D\mathbf z$, the induced probing law $\tilde q$ on $\mathbb R^D$ satisfies
\begin{equation}
\label{eq:subspace-reduction-integral}
\Delta_2^{(D)}(k+1)
=
\int_{\mathbb R^D}
\left(
a_k + \mathbf c_k^\top \mathbf z + \frac12 \mathbf z^\top \mathbf B_k \mathbf z
\right)^2
\, \tilde q(\mathbf z)\, d\mathbf z.
\end{equation}
\end{lemma}
The proof is given in Appendix~\ref{app:proofs}.

\paragraph{Main rate result.}
The main rate bound requires only local boundedness at the sequential minimizers.

\begin{assumption}[Uniform boundedness at sequential minimizers]
\label{ass:uniform-bounds}
There exist constants $M_\ell, M_{\mathbf g}, M_{\mathbf H}>0$, independent of $k$, such that for all $k\ge 1$ and all $i=1,\dots,k+1$,
\[
|\ell_i(\mathbf w_k^*)| \le M_\ell,
\qquad
\|\mathbf g_{k+1}(\mathbf w_k^*)\|_2 \le M_{\mathbf g},
\qquad
\|\mathbf H_i(\mathbf w_k^*)\|_2 \le M_{\mathbf H}.
\]
\end{assumption}
This assumption is used in the proof of Theorem~\ref{thm:subspace-rate}; see Appendix~\ref{app:proofs}.

\begin{theorem}[Subspace mean-squared rate]
\label{thm:subspace-rate}
Suppose Lemma~\ref{lem:subspace-reduction} and Assumption~\ref{ass:uniform-bounds} hold, and let $\tilde q(\mathbf z)=\mathcal N(\mathbf 0,\sigma^2\mathbf I_D)$. Then
\begin{equation}
\label{eq:subspace-rate-bound}
\Delta_2^{(D)}(k+1)
\le
\frac{
12M_\ell^2
+
3\sigma^2 M_{\mathbf g}^2
+
3\sigma^4(D^2+2D)M_{\mathbf H}^2
}{(k+1)^2}
=
\mathcal O(k^{-2}).
\end{equation}
\end{theorem}

The proof is given in Appendix~\ref{app:proofs}. Theorem~\ref{thm:subspace-rate} is a no-rate-loss statement under geometric compression. The criterion is evaluated on a $D$-dimensional curvature-aligned probe rather than under isotropic perturbations in $\mathbb R^N$, yet its mean-squared decay matches that of the full-space criterion.

\paragraph{Spectral interpretation.}
Under an additional stable-principal-directions regime, the compressed Hessian difference $\mathbf B_k$ becomes diagonal in the principal basis, and the criterion admits a closed form in terms of leading eigenvalue increments; see Corollary~\ref{cor:spectral} in Appendix~\ref{app:proofs}.

\paragraph{Extremality of the top-$D$ eigenspace.}
The following proposition formalizes the sense in which the top-$D$ choice is canonical within the eigenspace-aligned family.

\begin{proposition}[Extremality of the top-curvature subspace; informal]
\label{prop:extremal-topD}
Assume $a_k=0$, $\mathbf c_k=\mathbf 0$, and that $\mathbf H^{(k)}(\mathbf w_k^*)$ and $\mathbf H^{(k+1)}(\mathbf w_k^*)$ share a common eigenbasis with non-negative eigenvalue increments. Then among all $D$-dimensional eigenspace-aligned subspaces, the top-$D$ principal curvature subspace maximizes the pure quadratic stabilization signal $\Delta_{2,I}^{\mathrm{quad}}(k+1)$.
\end{proposition}

A precise statement and proof are given in Appendix~\ref{app:extremal}.

\section{Algorithmic estimation at scale}
\label{sec:algo}

Lemma~\ref{lem:subspace-reduction} and Eq.~\eqref{eq:subspace-reduction-integral} show that, once the probe is restricted to the principal curvature subspace, the criterion is determined by three compressed objects: the scalar value gap $a_k$, the projected gradient $\mathbf c_k\in\mathbb R^D$, and the compressed Hessian difference $\mathbf B_k\in\mathbb R^{D\times D}$.
This leads to a simple estimator taxonomy.
One estimator targets the true criterion directly.
Two cheaper estimators target its local quadratic surrogate.
All three share the same first step: construct the principal curvature subspace at $\mathbf w_k^*$.

\paragraph{Cost notation.}
Let $C_{\mathrm{fwd}}(m)$ denote the cost of one forward evaluation of $\mathcal L_m(\mathbf w)$, $C_{\mathrm{bwd}}(m)$ the cost of one backward pass, and $C_{\mathrm{HVP}}(m)$ the cost of one Hessian--vector product with $\mathbf H^{(m)}(\mathbf w)$.
We write $S$ for the Monte Carlo sample count, $D$ for the subspace dimension, and $T_{\mathrm{eig}}$ for the number of eigensolver iterations.

\paragraph{Shared step: principal curvature subspace.}
For each $k$, we compute the top-$D$ eigenvectors of $\mathbf H^{(k)}(\mathbf w_k^*)$ using deflated power iteration on Hessian--vector products; other HVP-based iterative eigensolvers such as Lanczos or LOBPCG apply identically. These follow the standard scalable second-order toolkit initiated by Pearlmutter and used in modern Hessian-analysis frameworks such as PyHessian~\citep{pearlmutter1994fast,yao2020pyhessian}. If each iteration requires $\mathcal O(D)$ Hessian--vector products, the one-time subspace construction cost is $\mathcal O \bigl(T_{\mathrm{eig}}\,D\,C_{\mathrm{HVP}}(k)\bigr)$. This cost is shared by all subspace-based estimators below.

\paragraph{Direct Monte Carlo for the true criterion.}
The most faithful estimator samples directly from the subspace probe: $\mathbf z_s \sim \mathcal N(\mathbf 0,\sigma^2\mathbf I_D)$, $\mathbf w_s=\mathbf w_k^*+\mathbf U_D\mathbf z_s$, $s=1,\dots,S$. This gives
\begin{equation}
\label{eq:direct-subspace-mc}
\widehat{\Delta}_{2,\mathrm{dir}}^{(D)}(k+1)
=
\frac{1}{S}\sum_{s=1}^S
\Bigl(
\mathcal L_{k+1}(\mathbf w_s)-\mathcal L_k(\mathbf w_s)
\Bigr)^2.
\end{equation}
It estimates the true criterion $\Delta_2^{(D)}$ and does not rely on the quadratic approximation. Its post-subspace cost is $\mathcal O \left( S\bigl( ND + C_{\mathrm{fwd}}(k)+C_{\mathrm{fwd}}(k+1) \bigr) \right)$, typically dominated by the two forward evaluations per sample.

\paragraph{Quadratic surrogate and its coefficients.}
Under the local quadratic model from Section~\ref{sec:theory}, the increment restricted to the principal curvature subspace is approximated by $a_k + \mathbf c_k^\top \mathbf z + \frac12 \mathbf z^\top \mathbf B_k \mathbf z$, where $a_k = \mathcal L_{k+1}(\mathbf w_k^*)-\mathcal L_k(\mathbf w_k^*)$, $\mathbf c_k = \mathbf U_D^\top \mathbf g^{(k+1)}(\mathbf w_k^*)$, and $\mathbf B_k = \mathbf U_D^\top \bigl( \mathbf H^{(k+1)}(\mathbf w_k^*)-\mathbf H^{(k)}(\mathbf w_k^*) \bigr) \mathbf U_D$. Assembling these coefficients requires one additional setup step with cost
\begin{equation}
\label{eq:proxy-setup-cost}
\mathcal O \left(
C_{\mathrm{fwd}}(k)+C_{\mathrm{fwd}}(k+1)
+
C_{\mathrm{bwd}}(k+1)
+
D(C_{\mathrm{HVP}}(k)+C_{\mathrm{HVP}}(k+1))
+
ND^2
\right).
\end{equation}

\paragraph{Quadratic Monte Carlo.}
Replacing the true increment in \eqref{eq:direct-subspace-mc} by the quadratic surrogate yields
\begin{equation}
\label{eq:quad-subspace-mc}
\widehat{\Delta}_{2,\mathrm{quadMC}}^{(D)}(k+1)
=
\frac{1}{S}\sum_{s=1}^S
\left(
a_k + \mathbf c_k^\top \mathbf z_s + \frac12 \mathbf z_s^\top \mathbf B_k \mathbf z_s
\right)^2,
\qquad
\mathbf z_s \sim \mathcal N(\mathbf 0,\sigma^2\mathbf I_D).
\end{equation}
After the coefficient setup in \eqref{eq:proxy-setup-cost}, its evaluation cost is $\mathcal O(SD^2)$.

\paragraph{Gaussian-moment estimator.}
Because the surrogate is quadratic in $\mathbf z$ and the probe is Gaussian, its expectation can be evaluated in closed form:
\begin{equation}
\label{eq:quad-closed-form}
\widehat{\Delta}_{2,\mathrm{GM}}^{(D)}(k+1)
=
\mathbb E_{\mathbf z\sim\mathcal N(\mathbf 0, \sigma^2 \mathbf I_D)}
\left[
\left(
a_k + \mathbf c_k^\top \mathbf z + \frac12 \mathbf z^\top \mathbf B_k \mathbf z
\right)^2
\right].
\end{equation}
Using Gaussian moment identities yields
\begin{equation}
\label{eq:quad-closed-form-expanded}
\widehat{\Delta}_{2,\mathrm{GM}}^{(D)}(k+1)
=
a_k^2
+
a_k \sigma^2 \operatorname{Tr}(\mathbf B_k)
+
\sigma^2 \|\mathbf c_k\|_2^2
+
\frac{\sigma^4}{4}
\left(
2\operatorname{Tr}(\mathbf B_k^2)
+
\operatorname{Tr}(\mathbf B_k)^2
\right).
\end{equation}
After setup, this estimator costs $\mathcal O(D^2)$.

\paragraph{Summary.}
The three estimators differ only in where they sit on the fidelity--efficiency spectrum.
Direct subspace Monte Carlo targets the true criterion and is therefore the reference estimator for $\Delta_2^{(D)}$.
Quadratic Monte Carlo and the Gaussian-moment estimator are much cheaper after setup, but they target only the local quadratic surrogate.
Their empirical comparison therefore tests two things at once: computational savings and the practical validity of the quadratic approximation.

\begin{table}[t]
\caption{Subspace-based estimators for $\Delta_2^{(D)}$. All methods share the one-time subspace construction cost $\mathcal O(T_{\mathrm{eig}}\,D\,C_{\mathrm{HVP}}(k))$. The most efficient one is Gaussian-moment (GM) estimator.}
\label{tab:estimator-summary}
\begin{center}
\small
\begin{tabular}{p{1.8cm} p{2.4cm} p{2.1cm} p{5.0cm}}
\toprule
\textbf{Estimator} & \textbf{Target} & \textbf{One-time setup} & \textbf{Evaluation cost} \\
\midrule
Direct MC
&
true $\Delta_2^{(D)}$
&
none
&
$\mathcal O \left(S\bigl(ND + C_{\mathrm{fwd}}(k)+C_{\mathrm{fwd}}(k+1)\bigr)\right)$
\\[0.45em]

Quadratic MC
&
quadratic surrogate
&
Eq.~\eqref{eq:proxy-setup-cost}
&
$\mathcal O(SD^2)$
\\[0.45em]

GM
&
quadratic surrogate
&
Eq.~\eqref{eq:proxy-setup-cost}
&
$\mathcal O(D^2)$
\\
\bottomrule
\end{tabular}
\end{center}
\vspace{-1.4\baselineskip}
\end{table}

\section{Experiments}\label{sec:experiments}

We evaluate four questions\footnote{Code for reproducing all figures and tables is available at \url{https://github.com/kisnikser/curvature-subspace-landscape}.}: how the proposed criteria decay with effective sample size, when the subspace criterion preserves the full-space mean-squared signal, for which perturbation scales the quadratic proxy is accurate, and how the three estimators trade fidelity against efficiency. Surprisingly, we observe that (i) a curvature-aligned probe occupying less than one part in a million of parameter space ($D/N < 10^{-6}$) already reproduces the full-space mean-squared signal to within numerical noise throughout the validated local regime, and (ii) once the subspace has been constructed, the closed-form Gaussian-moment estimator is roughly $18{,}000\times$ faster than direct Monte Carlo without measurable loss of fidelity.

\paragraph{Setup.}
All experiments use the \texttt{nanochat} depth-6 model (tag~\texttt{d6}), a 107M-parameter decoder-only transformer with rotary position embeddings, grouped-query attention, $\mathrm{ReLU}$ activations, and RMSNorm without learnable parameters, evaluated at training step~3500. We use this model because it is small enough to make repeated full-model Hessian--vector products tractable in float32 precision while still retaining nontrivial second-order structure. Autocast is disabled throughout, and we use the SDPA math kernel for numerical stability. Throughout this section, $k$ denotes the number of training sequences defining $\mathcal L_k$. Unless stated otherwise, we define $\mathcal L_k$ and $\mathcal L_{k+1}$ from $k=8$ sequences and construct the principal curvature subspace at $\mathbf w_k^*$ by deflated power iteration on Hessian--vector products; other HVP-based iterative eigensolvers such as Lanczos or LOBPCG apply identically.

\begin{wrapfigure}{r}{0.50\columnwidth}
    \vspace{-1.6\baselineskip}
    \centering
    \includegraphics[width=\linewidth]{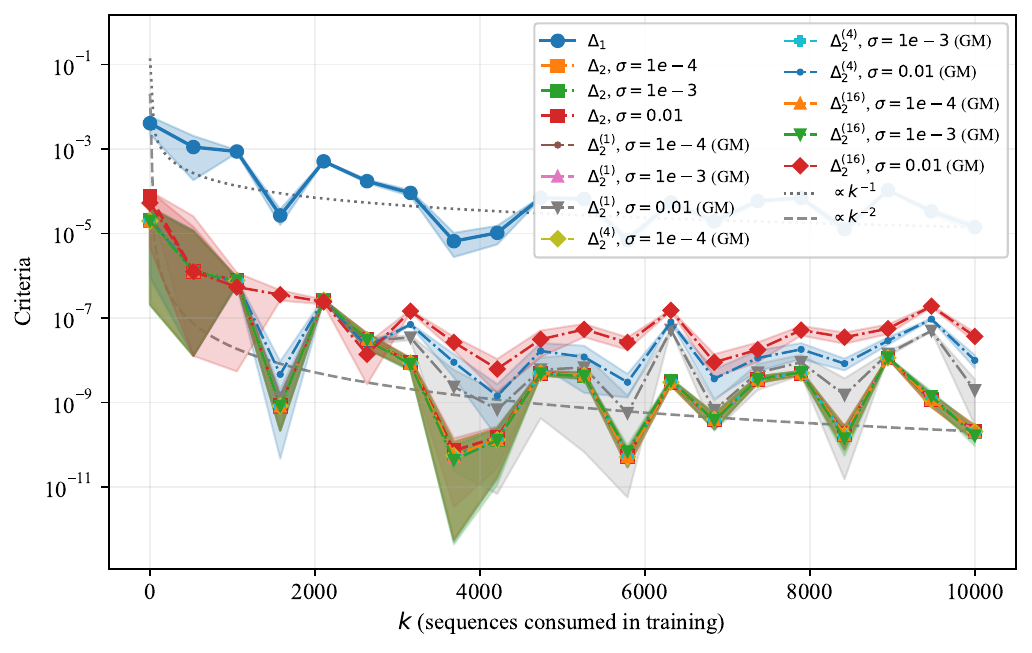}
    \caption{\textbf{Decay of stabilization criteria with sample size.} Comparison of $\Delta_1$, $\Delta_2$, and $\Delta_2^{(D)}$ as functions of $k$.}
    \label{fig:exp-k-decay}
    \vspace{-1.4\baselineskip}
\end{wrapfigure}

\paragraph{Criterion decay under sample growth.}
We first test whether the criteria from Sections~\ref{sec:method}--\ref{sec:theory} exhibit the predicted stabilization trend as the effective sample size grows. To do so, we evaluate the pointwise criterion $\Delta_1$, the isotropic mean-squared criterion $\Delta_2$, and the curvature-aware subspace criterion $\Delta_2^{(D)}$ across a range of sample sizes $k$.

Figure~\ref{fig:exp-k-decay} shows that all criteria decrease with $k$, but at different rates. The pointwise criterion decays more slowly, whereas the mean-squared criteria are orders of magnitude smaller and follow the steeper trend suggested by the quadratic scaling argument. The subspace criteria remain close in scale to the full-space mean-squared criterion while restricting the probe to a much smaller set of directions.

\begin{wrapfigure}{l}{0.50\columnwidth}
    \vspace{-2.0\baselineskip}
    \centering
    \includegraphics[width=\linewidth]{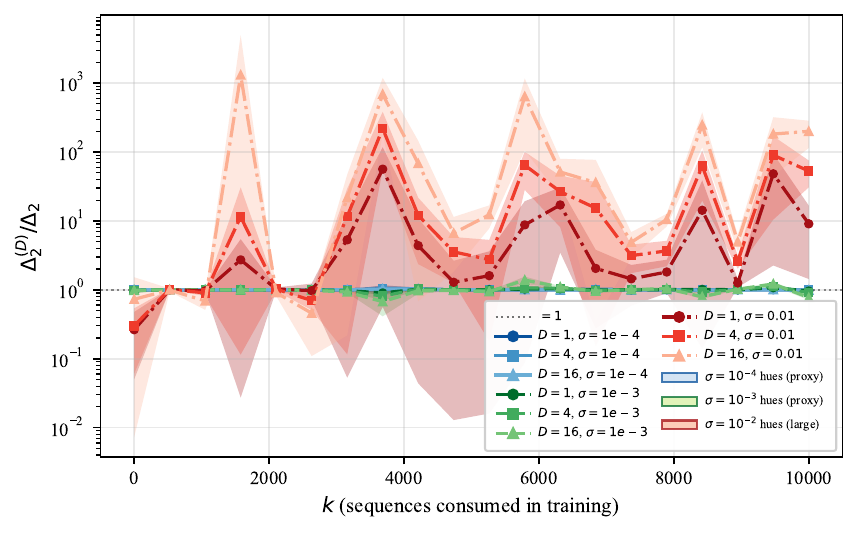}
    \caption{\textbf{Subspace criterion relative to the full-space criterion.} Ratio $\Delta_2^{(D)}/\Delta_2$ across sample size $k$, for several dimensions $D$ and scales $\sigma$.}
    \label{fig:exp-subspace-compare}
    \vspace{-3.0\baselineskip}
\end{wrapfigure}

\paragraph{Subspace versus full-space criterion.}
We next test whether the curvature-aligned subspace criterion preserves the full-space mean-squared signal. For several values of $D$ and $\sigma$, we track the ratio $\Delta_2^{(D)}/\Delta_2$ as a function of $k$.

Figure~\ref{fig:exp-subspace-compare} reveals a clear regime split. For $\sigma=10^{-4}$ and $\sigma=10^{-3}$, the ratio stays close to $1$ across the full range of $k$, indicating that curvature-aligned probing preserves essentially the same stabilization signal as the full-space criterion. For $\sigma=10^{-2}$, the ratio departs strongly from $1$, becomes much more variable, and depends clearly on subspace dimension, with larger $D$ producing systematically larger values. This behavior is consistent with leaving the local regime in which subspace compression remains faithful to the full-space observable.

\paragraph{Quadratic proxy validity.}
The theory and the proxy estimators of Section~\ref{sec:algo} rely on the local quadratic model in Eqs.~\eqref{eq:taylor-quadratic}--\eqref{eq:loss-diff-minimum}. Before using these estimators, we therefore identify the perturbation scales for which the approximation is accurate. For a fixed checkpoint $\mathbf w_k^*$, we draw isotropic Gaussian perturbations $\boldsymbol{\delta}\sim\mathcal{N}(\mathbf{0},\sigma^2\mathbf{I}_N)$ and compare the true local loss increment $\mathcal{L}_k(\mathbf{w}_k^*+\boldsymbol{\delta})-\mathcal{L}_k(\mathbf{w}_k^*)$ to its second-order Taylor approximation $\mathbf{g}^{(k)\top}\boldsymbol{\delta}+\frac{1}{2}\boldsymbol{\delta}^\top\mathbf{H}^{(k)}\boldsymbol{\delta}$.

\begin{wrapfigure}{r}{0.5\columnwidth}
    \vspace{-1.4\baselineskip}
    \centering
    \includegraphics[width=\linewidth]{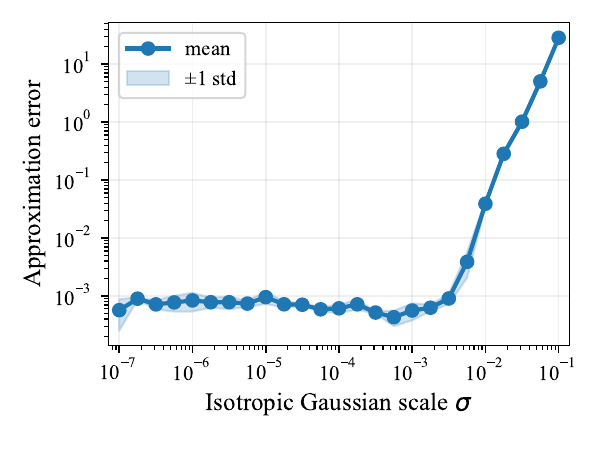}
    \caption{Relative error of the quadratic Taylor approximation versus perturbation scale $\sigma$ (mean and standard deviation across seeds, nanochat \texttt{d6}, step~3500).}
    \label{fig:exp-1}
    \vspace{-2.0\baselineskip}
\end{wrapfigure}

Figure~\ref{fig:exp-1} shows a clear local regime: the approximation error remains low and nearly flat up to about $\sigma\approx10^{-3}$, and then rises sharply for larger perturbations. We therefore restrict the proxy-based estimators to $\sigma\le10^{-3}$ in the remaining experiments.

\paragraph{Estimator fidelity and computational trade-offs.}
We finally compare the three estimators from Section~\ref{sec:algo}: direct subspace Monte Carlo, quadratic Monte Carlo, and the Gaussian-moment (GM) estimator. Direct~MC targets the true criterion $\Delta_2^{(D)}$, whereas Quadratic~MC and GM target its local quadratic surrogate.

\begin{wrapfigure}{l}{0.4\columnwidth}
    \vspace{-0.6\baselineskip}
    \centering
    \includegraphics[width=\linewidth]{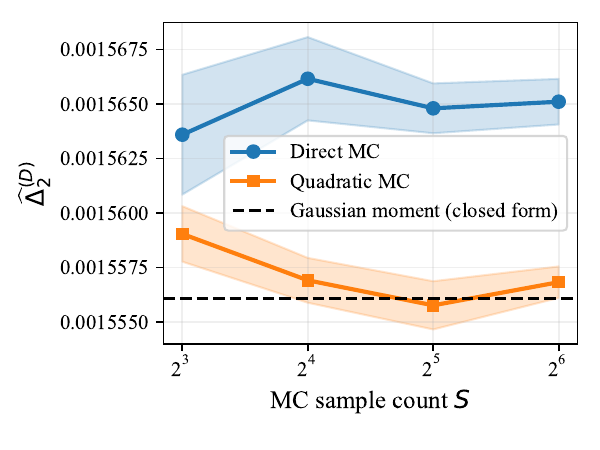}
    \caption{Convergence of Direct~MC and Quadratic~MC estimates of $\widehat{\Delta}_2^{(D)}$ to the Gaussian-moment closed form as the sample count $S$ increases, at $D=10$ and $\sigma=10^{-3}$.}
    \label{fig:exp-3}
    \vspace{-3.0\baselineskip}
\end{wrapfigure}

\begin{wrapfigure}{r}{0.55\columnwidth}
    \vspace{-0.6\baselineskip}
    \centering
    \includegraphics[width=\linewidth]{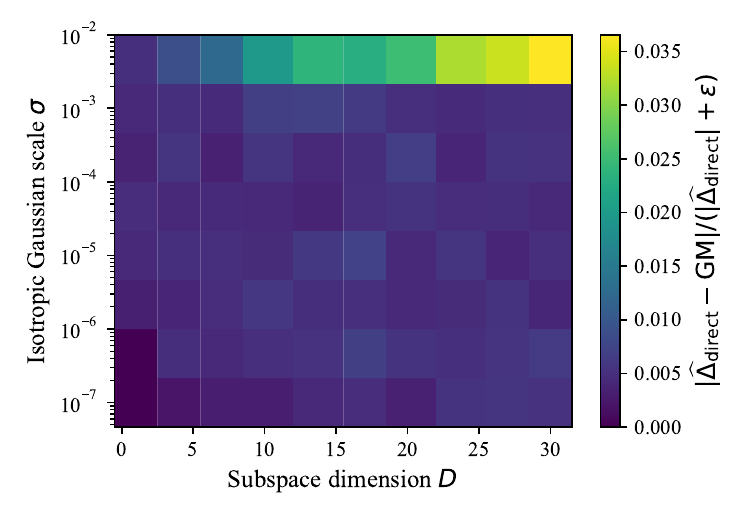}
    \caption{Relative error $|\widehat{\Delta}_{\mathrm{direct}}-\mathrm{GM}|/(|\widehat{\Delta}_{\mathrm{direct}}|+\varepsilon)$ over subspace dimension $D$ and perturbation scale $\sigma$.}
    \label{fig:exp-4}
    \vspace{-1.4\baselineskip}
\end{wrapfigure}

Figure~\ref{fig:exp-3} shows that both Monte Carlo estimators converge toward the GM value as $S$ increases. Figure~\ref{fig:exp-4} shows that the discrepancy between Direct~MC and GM stays small throughout the validated local regime, but grows at the largest tested perturbation scale.

\begin{table}[t]
\caption{Wall-clock times (seconds; mean $\pm$ sample standard deviation over five seeds) for the three estimators of $\Delta_2^{(D)}$ on the nanochat \texttt{d6} model ($D=45$, step~3500). Stage~I: top-$D$ Hessian eigenvectors (shared). Stage~II: gradient and compressed Hessian assembly (proxy estimators only). Stage~III: criterion evaluation.}
\label{tab:exp-2}
\begin{center}
\small
\begin{tabular}{lccc}
\toprule
\textbf{Estimator} & \textbf{Stage I (s)} & \textbf{Stage II (s)} & \textbf{Stage III (s)} \\
\midrule
Direct MC        & $17.53 \pm 0.53$ & \textemdash{}       & $2.197 \pm 0.013$ \\
Quadratic MC     & $17.53 \pm 0.53$ & $1.777 \pm 0.014$   & $(3.68 \pm 0.05)\times10^{-3}$ \\
GM               & $17.53 \pm 0.53$ & $1.777 \pm 0.014$   & $(1.22 \pm 0.03)\times10^{-4}$ \\
\bottomrule
\end{tabular}
\end{center}
\vspace{-1.6\baselineskip}
\end{table}

Table~\ref{tab:exp-2} shows that the shared subspace-construction stage dominates the total runtime. Once that cost has been paid, GM is by far the cheapest estimator: its evaluation stage is about $18{,}000\times$ faster than Direct~MC. Taken together, these results suggest a simple practical picture: Direct~MC is the reference estimator for the true criterion, but in the validated local regime the GM proxy provides nearly the same signal at negligible additional cost.

\section{Limitations}
\label{sec:limitations}

Our analysis is inherently local. The main theoretical result is derived under a second-order approximation of the increment field near a trained solution $\mathbf w_k^*$, so its interpretation is restricted to perturbation regimes in which that approximation is accurate. Accordingly, the proxy estimators from Section~\ref{sec:algo} should be viewed as local surrogates rather than globally faithful approximations. Theorem~\ref{thm:subspace-rate} is a one-sided rate statement: it certifies that geometric compression preserves the $\mathcal O(k^{-2})$ decay, but we do not claim tightness of the $(k+1)^{-2}$ bound or of its $\sigma^4(D^2+2D)$ dependence, and the constant carries a $D^2$ term that trades against the ambient-dimension factor only for moderate $D$.

The top-$D$ principal-curvature subspace is motivated by empirical anisotropy and supported by Proposition~\ref{prop:extremal-topD}, but this extremality result holds only in an idealized simultaneously-diagonalizable quadratic regime. More general non-quadratic or rapidly drifting regimes may favor adaptive subspaces, and Assumption~\ref{ass:uniform-bounds} is itself a local statement whose plausibility near sequential minimizers we do not prove beyond boundedness.

Our empirical study is intentionally narrow. We focus on a 107M-parameter decoder-only transformer because repeated full-model second-order computations are still feasible in float32. Additional ablations on other architectures and sizes showed a similar qualitative picture, but are omitted for brevity. This scope does not guarantee that the same geometric effects or computational trade-offs transfer unchanged to much larger models or different training regimes. Finally, although Hessian--vector products and iterative eigensolvers are far cheaper than explicit Hessian construction, they still add nontrivial overhead, with subspace construction dominating the cost in our setting.

\section{Discussion}
\label{sec:discussion}

Our main claim is conceptual as much as technical: under sample growth, stabilization depends not only on which functional of the increment field is aggregated, but also on how the field is probed. The proposed criterion is therefore not just a lower-dimensional version of an existing mean-squared observable; it makes the probing law explicit and aligns it with the anisotropic local geometry of the empirical landscape.

Within the local quadratic regime, this geometric restriction preserves the canonical $\mathcal O(k^{-2})$ mean-squared decay while replacing ambient-dimension curvature dependence by dependence on the probing dimension $D$. Empirically, the subspace criterion tracks the full-space mean-squared signal throughout the validated local regime, and the Gaussian-moment proxy reproduces the same signal much faster than direct Monte Carlo once the subspace is available.

More broadly, curvature-aligned probing is one instance of a larger family of geometry-aware local observables. If stabilization is viewed as an observational problem, other structured probing laws may reveal aspects of landscape deformation under data growth that curvature alone does not capture. For example, data-dependent subspaces spanned by gradients of influential examples fit the same framework and suggest a broader class of geometry- and data-aware observables compatible with the estimator machinery developed here.

\section{Conclusion}
\label{sec:conclusion}

We introduced a unified view of local loss-landscape stabilization and proposed a curvature-aligned subspace criterion based on the top-$D$ Hessian eigenspace near a trained solution. Under a local quadratic model, this criterion preserves the full-space mean-squared $\mathcal O(k^{-2})$ rate with dependence on the probing dimension $D$, and admits scalable estimators that are efficient in the valid local regime. This enables a concrete quantitative question---how much of one-sample landscape deformation is concentrated in the dominant curvature modes---to be answered with a closed-form observable. We do not claim that the top-$D$ eigenspace is universally optimal beyond the eigenspace-aligned quadratic regime, nor that a single $\mathcal O(k^{-2})$ bound settles the sample-growth behavior of modern training pipelines; both are natural directions for future work.


\bibliographystyle{unsrtnat}
\bibliography{references}


\appendix

\section{Proofs}
\label{app:proofs}

\allowdisplaybreaks

\subsection{Spectral interpretation under stable principal directions}
\label{app:spectral}

The rate bound in Theorem~\ref{thm:subspace-rate} does not require any alignment between the eigenspaces of $\mathbf H^{(k)}(\mathbf w_k^*)$ and $\mathbf H^{(k+1)}(\mathbf w_k^*)$. For interpretation, it is useful to isolate a more structured regime in which the leading curvature directions remain stable across the one-sample increment.

\begin{assumption}[Stable principal directions]
\label{ass:stable-principal}
The eigenvectors $\mathbf u_1,\dots,\mathbf u_D$ associated with the $D$ largest eigenvalues of $\mathbf H^{(k)}(\mathbf w_k^*)$ are also eigenvectors of $\mathbf H^{(k+1)}(\mathbf w_k^*)$:
\[
\mathbf H^{(k+1)}(\mathbf w_k^*)\,\mathbf u_i
=
\lambda_i^{(k+1)}\mathbf u_i,
\qquad i=1,\dots,D.
\]
\end{assumption}

Under Assumption~\ref{ass:stable-principal},
\[
\mathbf B_k
=
\operatorname{diag}\bigl(
\lambda_1^{(k+1)}-\lambda_1^{(k)},
\dots,
\lambda_D^{(k+1)}-\lambda_D^{(k)}
\bigr),
\]
so the compressed Hessian difference becomes diagonal in the principal basis.

\begin{corollary}[Spectral closed form under vanishing value and linear terms]
\label{cor:spectral}
Suppose Assumption~\ref{ass:stable-principal} holds, $\tilde q(\mathbf z)=\mathcal N(\mathbf 0,\sigma^2\mathbf I_D)$, and $a_k=0$, $\mathbf c_k=\mathbf 0$. Then
\begin{equation}
\label{eq:spectral-closed-form}
\Delta_2^{(D)}(k+1)
=
\frac{\sigma^4}{4}
\left(
2\sum_{i=1}^D (\lambda_i^{(k+1)}-\lambda_i^{(k)})^2
+
\left(
\sum_{i=1}^D (\lambda_i^{(k+1)}-\lambda_i^{(k)})
\right)^2
\right).
\end{equation}
\end{corollary}

Corollary~\ref{cor:spectral} isolates the pure quadratic regime. The assumptions $a_k=0$ and $\mathbf c_k=\mathbf 0$ are idealizations, but they become increasingly natural when the value gap is small and $\mathbf w_k^*$ lies close to a minimizer of $\mathcal L_{k+1}$ as well. In that regime, the criterion becomes a direct function of the leading eigenvalue increments.

\subsection{Lemma: Extremality of the top-curvature subspace}
\label{app:extremal}

\begin{lemma}[Extremality of the top-curvature subspace]
\label{lemma:extremal-topD}
Assume that $a_k=0$, $\mathbf c_k=\mathbf 0$.
Suppose that $\mathbf H^{(k)}(\mathbf w_k^*)$ and $\mathbf H^{(k+1)}(\mathbf w_k^*)$ are simultaneously diagonalizable with a common orthonormal eigenbasis $\{\mathbf u_i\}_{i=1}^N$, and let $\delta_i=\lambda_i^{(k+1)}-\lambda_i^{(k)}$, $i=1,\dots,N$.
Assume further that $\delta_1 \ge \delta_2 \ge \cdots \ge \delta_N \ge 0$.
For any index set $I\subset\{1,\dots,N\}$ with $|I|=D$, let $\mathcal S_I=\operatorname{span}\{\mathbf u_i:\ i\in I\}$, and define the corresponding quadratic subspace criterion by restricting \eqref{eq:delta2-subspace-method} to $\mathbf w_k^*+\mathcal S_I$ under the local quadratic model, with Gaussian coordinates $\mathbf z\sim\mathcal N(\mathbf 0,\sigma^2\mathbf I_D)$.
Then
\begin{equation}
\label{eq:extremal-criterion}
\Delta_{2,I}^{\mathrm{quad}}(k+1)
=
\frac{\sigma^4}{4}
\left(
2\sum_{i\in I}\delta_i^2
+
\left(\sum_{i\in I}\delta_i\right)^2
\right),
\end{equation}
and the maximum over all index sets $I$ with $|I|=D$ is attained at $I^*=\{1,\dots,D\}$.
Equivalently, the top-$D$ eigenspace of $\mathbf H^{(k)}(\mathbf w_k^*)$ maximizes the pure quadratic stabilization signal within the eigenspace-aligned family.
\end{lemma}

Under the simultaneous diagonalizability assumption, $\mathbf B_k = \operatorname{diag}(\delta_1,\dots,\delta_N)$ restricted to $\mathcal S_I$ gives $\mathbf B_{k,I}=\operatorname{diag}(\delta_i : i\in I)$.
With $a_k=0$, $\mathbf c_k=\mathbf 0$, Lemma~\ref{lem:subspace-reduction} and the Gaussian moment identities of Theorem~\ref{thm:subspace-rate} yield \eqref{eq:extremal-criterion}.
The function
\[
f(I)=2\sum_{i\in I}\delta_i^2+\left(\sum_{i\in I}\delta_i\right)^2
\]
is maximized by choosing the $D$ largest $\delta_i$ values, i.e., $I^*=\{1,\dots,D\}$, since all terms are non-negative and $\delta_1\ge\cdots\ge\delta_N\ge 0$.

\subsection{Proof of Lemma~\ref{lem:subspace-reduction}}

Since the columns of $\mathbf U_D\in\mathbb R^{N\times D}$ are orthonormal, every point of $\mathbf w_k^*+\mathcal S_D$ has a unique representation
\[
\mathbf w=\mathbf w_k^*+\mathbf U_D\mathbf z,
\qquad
\mathbf z\in\mathbb R^D.
\]
Substituting into the local quadratic model \eqref{eq:loss-diff-minimum} and using $\mathbf g^{(k)}(\mathbf w_k^*)=\mathbf 0$ gives
\begin{align*}
\mathcal L_{k+1}(\mathbf w)-\mathcal L_k(\mathbf w)
&=
a_k
+
\mathbf g^{(k+1)}(\mathbf w_k^*)^\top \mathbf U_D\mathbf z
+
\frac12(\mathbf U_D\mathbf z)^\top \mathbf A_k (\mathbf U_D\mathbf z)
\\
&=
a_k + \mathbf c_k^\top \mathbf z + \frac12 \mathbf z^\top \mathbf B_k \mathbf z.
\end{align*}
Under the parameterization $\mathbf w=\mathbf w_k^*+\mathbf U_D\mathbf z$, the density $q$ supported on $\mathbf w_k^*+\mathcal S_D$ induces a density $\tilde q$ on $\mathbb R^D$. Integrating yields \eqref{eq:subspace-reduction-integral}.

\subsection{Proof of Theorem~\ref{thm:subspace-rate}}

By Lemma~\ref{lem:subspace-reduction},
\[
\Delta_2^{(D)}(k+1)
=
\mathbb E_{\tilde q(\mathbf z)}
\left[
\left(
a_k+\mathbf c_k^\top\mathbf z+\frac12\mathbf z^\top\mathbf B_k\mathbf z
\right)^2
\right],
\qquad
\mathbf z\sim \mathcal N(\mathbf 0,\sigma^2\mathbf I_D).
\]
For any $x,y,z\in\mathbb R$,
\[
(x+y+z)^2\le 3(x^2+y^2+z^2),
\]
hence
\begin{equation}
\label{eq:app-thm-step1-new}
\Delta_2^{(D)}(k+1)
\le
3a_k^2
+
3\,\mathbb E\bigl[(\mathbf c_k^\top\mathbf z)^2\bigr]
+
\frac34\,\mathbb E\bigl[(\mathbf z^\top\mathbf B_k\mathbf z)^2\bigr].
\end{equation}

\paragraph{Zero-order term.}
Using \eqref{eq:loss-increment-exact} at $\mathbf w_k^*$,
\[
a_k
=
\mathcal L_{k+1}(\mathbf w_k^*)-\mathcal L_k(\mathbf w_k^*)
=
\frac{1}{k+1}\Bigl(\ell_{k+1}(\mathbf w_k^*)-\mathcal L_k(\mathbf w_k^*)\Bigr).
\]
Therefore
\[
|a_k|
\le
\frac{1}{k+1}
\Bigl(
|\ell_{k+1}(\mathbf w_k^*)|
+
|\mathcal L_k(\mathbf w_k^*)|
\Bigr).
\]
Moreover,
\[
|\mathcal L_k(\mathbf w_k^*)|
=
\left|
\frac1k\sum_{i=1}^k \ell_i(\mathbf w_k^*)
\right|
\le
\frac1k\sum_{i=1}^k |\ell_i(\mathbf w_k^*)|
\le
M_\ell,
\]
and also $|\ell_{k+1}(\mathbf w_k^*)|\le M_\ell$. Hence
\begin{equation}
\label{eq:app-ak-sq-new}
|a_k|\le \frac{2M_\ell}{k+1},
\qquad
a_k^2\le \frac{4M_\ell^2}{(k+1)^2}.
\end{equation}

\paragraph{Linear term.}
Since $\mathbf g^{(k)}(\mathbf w_k^*)=\mathbf 0$,
\[
\mathbf b_k
=
\mathbf g^{(k+1)}(\mathbf w_k^*)
=
\frac{1}{k+1}\mathbf g_{k+1}(\mathbf w_k^*),
\qquad
\mathbf c_k=\mathbf U_D^\top\mathbf b_k.
\]
Because $\mathbf U_D$ has orthonormal columns,
\[
\|\mathbf c_k\|_2
\le
\|\mathbf U_D^\top\|_2\,\|\mathbf b_k\|_2
\le
\frac{M_{\mathbf g}}{k+1}.
\]
For $\mathbf z\sim\mathcal N(\mathbf 0,\sigma^2\mathbf I_D)$,
\begin{equation}
\label{eq:app-linear-new}
\mathbb E\bigl[(\mathbf c_k^\top\mathbf z)^2\bigr]
=
\mathbf c_k^\top \mathbb E[\mathbf z\mathbf z^\top]\mathbf c_k
=
\sigma^2\|\mathbf c_k\|_2^2
\le
\frac{\sigma^2 M_{\mathbf g}^2}{(k+1)^2}.
\end{equation}

\paragraph{Quadratic term.}
The matrix $\mathbf B_k$ is symmetric. For $\mathbf z\sim\mathcal N(\mathbf 0,\sigma^2\mathbf I_D)$,
\begin{equation}
\label{eq:app-quad-moment-new}
\mathbb E\bigl[(\mathbf z^\top\mathbf B_k\mathbf z)^2\bigr]
=
2\sigma^4 \operatorname{Tr}(\mathbf B_k^2)
+
\sigma^4 \operatorname{Tr}(\mathbf B_k)^2.
\end{equation}
For symmetric $\mathbf B_k$,
\[
\operatorname{Tr}(\mathbf B_k^2)\le D\|\mathbf B_k\|_2^2,
\qquad
|\operatorname{Tr}(\mathbf B_k)|\le D\|\mathbf B_k\|_2,
\]
so
\begin{equation}
\label{eq:app-quad-bound-new}
\mathbb E\bigl[(\mathbf z^\top\mathbf B_k\mathbf z)^2\bigr]
\le
\sigma^4(D^2+2D)\|\mathbf B_k\|_2^2.
\end{equation}

Next,
\[
\|\mathbf B_k\|_2
=
\|\mathbf U_D^\top \mathbf A_k \mathbf U_D\|_2
\le
\|\mathbf U_D^\top\|_2\,\|\mathbf A_k\|_2\,\|\mathbf U_D\|_2
\le
\|\mathbf A_k\|_2.
\]
Using
\[
\mathbf H^{(m)}(\mathbf w_k^*)
=
\frac1m\sum_{i=1}^m \mathbf H_i(\mathbf w_k^*),
\]
we obtain
\begin{align*}
\mathbf A_k
&=
\mathbf H^{(k+1)}(\mathbf w_k^*)-\mathbf H^{(k)}(\mathbf w_k^*)
\\
&=
\frac{1}{k+1}\sum_{i=1}^{k+1}\mathbf H_i(\mathbf w_k^*)
-
\frac{1}{k}\sum_{i=1}^{k}\mathbf H_i(\mathbf w_k^*)
\\
&=
\frac{1}{k+1}
\Bigl(
\mathbf H_{k+1}(\mathbf w_k^*)-\mathbf H^{(k)}(\mathbf w_k^*)
\Bigr).
\end{align*}
Hence
\[
\|\mathbf A_k\|_2
\le
\frac{1}{k+1}
\Bigl(
\|\mathbf H_{k+1}(\mathbf w_k^*)\|_2
+
\|\mathbf H^{(k)}(\mathbf w_k^*)\|_2
\Bigr)
\le
\frac{2M_{\mathbf H}}{k+1},
\]
and therefore
\begin{equation}
\label{eq:app-B-bound-new}
\|\mathbf B_k\|_2\le \frac{2M_{\mathbf H}}{k+1},
\qquad
\mathbb E\bigl[(\mathbf z^\top\mathbf B_k\mathbf z)^2\bigr]
\le
\frac{4\sigma^4(D^2+2D)M_{\mathbf H}^2}{(k+1)^2}.
\end{equation}

\paragraph{Conclusion.}
Substituting \eqref{eq:app-ak-sq-new}, \eqref{eq:app-linear-new}, and \eqref{eq:app-B-bound-new} into \eqref{eq:app-thm-step1-new}, we obtain
\[
\Delta_2^{(D)}(k+1)
\le
\frac{
12M_\ell^2
+
3\sigma^2M_{\mathbf g}^2
+
3\sigma^4(D^2+2D)M_{\mathbf H}^2
}{(k+1)^2},
\]
which is exactly \eqref{eq:subspace-rate-bound}.

\subsection{Proof of Corollary~\ref{cor:spectral}}

Under the assumptions $a_k=0$ and $\mathbf c_k=\mathbf 0$, Lemma~\ref{lem:subspace-reduction} gives
\[
\Delta_2^{(D)}(k+1)
=
\frac14\,
\mathbb E\bigl[(\mathbf z^\top\mathbf B_k\mathbf z)^2\bigr],
\qquad
\mathbf z\sim\mathcal N(\mathbf 0,\sigma^2\mathbf I_D).
\]
Using \eqref{eq:app-quad-moment-new},
\[
\mathbb E\bigl[(\mathbf z^\top\mathbf B_k\mathbf z)^2\bigr]
=
2\sigma^4\operatorname{Tr}(\mathbf B_k^2)
+
\sigma^4\operatorname{Tr}(\mathbf B_k)^2.
\]
Under Assumption~\ref{ass:stable-principal},
\[
\mathbf B_k
=
\operatorname{diag}\bigl(
\lambda_1^{(k+1)}-\lambda_1^{(k)},
\dots,
\lambda_D^{(k+1)}-\lambda_D^{(k)}
\bigr).
\]
Therefore,
\[
\operatorname{Tr}(\mathbf B_k)
=
\sum_{i=1}^D \bigl(\lambda_i^{(k+1)}-\lambda_i^{(k)}\bigr),
\qquad
\operatorname{Tr}(\mathbf B_k^2)
=
\sum_{i=1}^D \bigl(\lambda_i^{(k+1)}-\lambda_i^{(k)}\bigr)^2.
\]
Substituting these identities into the previous display yields
\[
\Delta_2^{(D)}(k+1)
=
\frac{\sigma^4}{4}
\left(
2\sum_{i=1}^D (\lambda_i^{(k+1)}-\lambda_i^{(k)})^2
+
\left(
\sum_{i=1}^D (\lambda_i^{(k+1)}-\lambda_i^{(k)})
\right)^2
\right),
\]
which is exactly \eqref{eq:spectral-closed-form}.

\subsection{Proof of Lemma~\ref{lemma:extremal-topD}}

Fix an index set $I\subset\{1,\dots,N\}$ with $|I|=D$ and consider the eigenspace-aligned subspace
\[
\mathcal S_I=\operatorname{span}\{\mathbf u_i:\ i\in I\}.
\]
Under the assumptions of the lemma, the linear and value terms vanish, and the Hessian difference is diagonal in the common eigenbasis:
\[
\mathbf A_k
=
\mathbf H^{(k+1)}(\mathbf w_k^*)-\mathbf H^{(k)}(\mathbf w_k^*)
=
\sum_{i=1}^N \delta_i\, \mathbf u_i\mathbf u_i^\top,
\qquad
\delta_i=\lambda_i^{(k+1)}-\lambda_i^{(k)}.
\]
Let $\mathbf U_I$ denote the matrix whose columns are the vectors $\mathbf u_i$, $i\in I$.
Then the compressed Hessian difference on $\mathcal S_I$ is
\[
\mathbf B_{k,I}
=
\mathbf U_I^\top \mathbf A_k \mathbf U_I
=
\operatorname{diag}(\delta_i:\ i\in I).
\]
With Gaussian coordinates $\mathbf z\sim\mathcal N(\mathbf 0,\sigma^2\mathbf I_D)$, the local quadratic reduction (same argument as in Lemma~\ref{lem:subspace-reduction}, with $\mathbf U_D$ replaced by $\mathbf U_I$) and $a_k=0$, $\mathbf c_k=\mathbf 0$ give
\[
\Delta_{2,I}^{\mathrm{quad}}(k+1)
=
\frac14\,\mathbb E\bigl[(\mathbf z^\top \mathbf B_{k,I}\mathbf z)^2\bigr].
\]
Applying the Gaussian quadratic-form identity \eqref{eq:app-quad-moment-new},
\[
\mathbb E\bigl[(\mathbf z^\top \mathbf B_{k,I}\mathbf z)^2\bigr]
=
2\sigma^4 \operatorname{Tr}(\mathbf B_{k,I}^2)
+
\sigma^4 \operatorname{Tr}(\mathbf B_{k,I})^2.
\]
Since $\mathbf B_{k,I}$ is diagonal,
\[
\operatorname{Tr}(\mathbf B_{k,I}^2)=\sum_{i\in I}\delta_i^2,
\qquad
\operatorname{Tr}(\mathbf B_{k,I})=\sum_{i\in I}\delta_i.
\]
Therefore
\[
\Delta_{2,I}^{\mathrm{quad}}(k+1)
=
\frac{\sigma^4}{4}
\left(
2\sum_{i\in I}\delta_i^2
+
\left(\sum_{i\in I}\delta_i\right)^2
\right),
\]
which proves \eqref{eq:extremal-criterion}.

It remains to show that this quantity is maximized when $I=\{1,\dots,D\}$.
Define
\[
F(I)
=
2\sum_{i\in I}\delta_i^2
+
\left(\sum_{i\in I}\delta_i\right)^2.
\]
Since $\delta_1\ge\delta_2\ge\cdots\ge\delta_N\ge0$, both terms in $F(I)$ are monotone with respect to replacing a smaller selected $\delta_j$ by a larger unselected $\delta_i$. More explicitly, suppose $i\notin I$, $j\in I$, and $\delta_i\ge \delta_j$, and define
\[
I'=(I\setminus\{j\})\cup\{i\}.
\]
Then
\[
\sum_{\ell\in I'}\delta_\ell
=
\sum_{\ell\in I}\delta_\ell - \delta_j + \delta_i
\ge
\sum_{\ell\in I}\delta_\ell,
\]
and similarly
\[
\sum_{\ell\in I'}\delta_\ell^2
=
\sum_{\ell\in I}\delta_\ell^2 - \delta_j^2 + \delta_i^2
\ge
\sum_{\ell\in I}\delta_\ell^2.
\]
Hence
\[
F(I')\ge F(I).
\]
By repeatedly exchanging smaller selected indices for larger unselected ones, one reaches the set $I^*=\{1,\dots,D\}$ without decreasing $F$. Therefore $F(I)$, and hence $\Delta_{2,I}^{\mathrm{quad}}(k+1)$, is maximized at $I^*=\{1,\dots,D\}$. This proves the lemma.

\section{Additional experimental details}
\label{app:exp}

\paragraph{Model and numerical setup.}
All main-text experiments use the \texttt{nanochat} depth-6 model (tag \texttt{d6}) evaluated at training step~3500. The model is a decoder-only transformer with rotary position embeddings, grouped-query attention, $\mathrm{ReLU}$ activations, and RMSNorm without learnable parameters. We use this setup because it is small enough to make repeated full-model Hessian--vector products feasible in float32 precision while still retaining nontrivial transformer loss geometry. Autocast is disabled throughout, and the SDPA math kernel is used for numerical stability.

\paragraph{Losses, checkpoints, and local quantities.}
Unless stated otherwise, the local criteria are evaluated at a checkpoint denoted by $\mathbf w_k^*$. The empirical risks $\mathcal L_k$ and $\mathcal L_{k+1}$ are formed from the corresponding nested sequence subsets used in the experiment under consideration. The local quadratic quantities in Sections~\ref{sec:theory} and~\ref{sec:algo} are computed at the same checkpoint:
\[
a_k = \mathcal L_{k+1}(\mathbf w_k^*)-\mathcal L_k(\mathbf w_k^*),\qquad
\mathbf c_k = \mathbf U_D^\top \mathbf g^{(k+1)}(\mathbf w_k^*),\qquad
\mathbf B_k=\mathbf U_D^\top\bigl(\mathbf H^{(k+1)}-\mathbf H^{(k)}\bigr)\mathbf U_D.
\]

\paragraph{Subspace construction.}
Top-$D$ Hessian eigendirections are computed from $\mathbf H^{(k)}(\mathbf w_k^*)$ using Hessian--vector products and iterative eigensolvers, as described in Section~\ref{sec:algo}. Unless stated otherwise, the principal curvature subspace is recomputed for each evaluated checkpoint.

\paragraph{Criterion estimation.}
For $\Delta_1$, we evaluate the one-point increment directly at $\mathbf w_k^*$. For $\Delta_2$ and $\Delta_2^{(D)}$, Monte Carlo estimates use Gaussian perturbations with the perturbation scales and sample counts reported in the corresponding figure captions. Direct subspace Monte Carlo samples perturbations in the principal curvature subspace, while Quadratic~MC and the Gaussian-moment estimator use the compressed surrogate coefficients $(a_k,\mathbf c_k,\mathbf B_k)$. Unless stated otherwise, proxy-based estimators are used only in the perturbation regime where the quadratic approximation is empirically validated.

\paragraph{Quadratic proxy validation.}
For the proxy-validity experiment in the main text, we draw isotropic Gaussian perturbations
\[
\boldsymbol{\delta}\sim\mathcal N(\mathbf 0,\sigma^2\mathbf I_N)
\]
and compare the true local loss increment
\[
\mathcal L_k(\mathbf w_k^*+\boldsymbol{\delta})-\mathcal L_k(\mathbf w_k^*)
\]
to its second-order Taylor approximation
\[
\mathbf g^{(k)\top}\boldsymbol{\delta}+\frac12\boldsymbol{\delta}^\top \mathbf H^{(k)}\boldsymbol{\delta}.
\]
The reported relative errors are averaged over random seeds and perturbation samples.

\paragraph{Decay under sample growth.}
For the sample-growth experiment in the main text, we evaluate $\Delta_1$, $\Delta_2$, and $\Delta_2^{(D)}$ across a grid of effective sample sizes $k$. The corresponding curves are intended to test the predicted decay of the stabilization criteria and to compare pointwise, isotropic, and curvature-aligned probes on the same training trajectory. When the experiment uses multiple random seeds, the plotted values are averaged across seeds and displayed with the uncertainty convention specified in the figure caption.

\paragraph{Runtime measurements.}
Wall-clock times in Table~\ref{tab:exp-2} are measured separately for three stages: subspace construction, surrogate-coefficient assembly, and criterion evaluation. The first stage is shared by all estimators. Reported values are averaged over five random seeds and shown as mean $\pm$ sample standard deviation.

\paragraph{Recorded outputs.}
For each run, we record the estimated criteria, proxy-validation diagnostics, subspace-comparison results, and estimator timing statistics needed to generate the figures and tables in the main text.

\end{document}